%% file: main.tex
\PassOptionsToPackage{dvipsnames,table}{xcolor}  
\documentclass[10pt,twocolumn,letterpaper]{article}
\usepackage{cvpr}

\definecolor{cvprblue}{rgb}{0.21,0.49,0.74}

\usepackage{pifont,bbding,fontawesome}
\usepackage[normalem]{ulem}
\usepackage[pagebackref,breaklinks,colorlinks,allcolors=cvprblue]{hyperref}
\usepackage{array,tabularx}
\usepackage{algorithm,algorithmic}
\usepackage{amsmath,amssymb,booktabs}
\usepackage{graphicx}  
\usepackage{multirow,makecell}
\usepackage{enumitem,placeins}
\usepackage{fancyhdr}

\def\paperID{10313} 
\def\confName{CVPR}
\def\confYear{2025}

\title{SF${^2}$T: Self-supervised Fragment Finetuning of Video-LLMs for Fine-Grained Understanding}

\author{%
Yangliu Hu$^1$,
Zikai Song$^{1\dagger}$,
Na Feng$^1$,
Yawei Luo$^2$,
Junqing Yu$^1$,
Yi-Ping Phoebe Chen$^3$,
Wei Yang$^{1\dagger}$
\smallskip
\\
$^1$Huazhong University of Science and Technology~~~
$^2$Zhejiang University~~~
$^3$La Trobe University\\
{\tt\small \{huyangliu,skyesong,fengna,yjqing,weiyangcs\}@hust.edu.cn}\\
{\tt\small yaweiluo@zju.edu.cn}~~~
{\tt\small phoebe.chen@latrobe.edu.au}
}

\begin{document}
\maketitle
\renewcommand{\thefootnote}{}
\footnotetext{$\dagger$ Corresponding authors}
\input{0_abstract}   
\input{1_intro}
\input{2_relatedwork}
\input{4_benchmark}

\input{3_method}
\input{5_experiment}
\input{6_conclusion}

\section*{Acknowledgments}
This work is supported by the National Key Research and Development Program of China (No.2020YBF2901202), National Natural Science Foundation of China (NSFC No. 62272184 and No. 62402189), the China Postdoctoral Science Foundation under Grant Number GZC20230894, the China Postdoctoral Science Foundation (Certificate Number: 2024M751012), and the Postdoctor Project of Hubei Province under Grant Number 2024HBBHCXB014, and the ``Pioneer'' and ``Leading Goose'' R\&D Program of Zhejiang (No. 2024C01161). The computation is completed in the HPC Platform of Huazhong University of Science and Technology.

{
    \small
    \normalem\bibliographystyle{ieeenat_fullname}
    \normalem\bibliography{main}
}

\input{X_suppl}

\end{document}

%% file: 0_abstract.tex
\begin{abstract}
\label{sec:abs}

Video-based Large Language Models (Video-LLMs) have witnessed substantial advancements in recent years, propelled by the advancement in multi-modal LLMs. Although these models have demonstrated proficiency in providing the overall description of videos, they struggle with fine-grained understanding, particularly in aspects such as visual dynamics and video details inquiries. 
To tackle these shortcomings, we find that fine-tuning Video-LLMs on self-supervised fragment tasks, greatly improve their fine-grained video understanding abilities. Hence we propose two key contributions:
(1) Self-Supervised Fragment Fine-Tuning (SF$^2$T), a novel effortless fine-tuning method, employs the rich inherent characteristics of videos for training, while unlocking more fine-grained understanding ability of Video-LLMs. Moreover, it relieves 
researchers from labor-intensive annotations and smartly circumvents the limitations of natural language, which often fails to capture the complex spatiotemporal variations in videos;
(2) A novel benchmark dataset, namely FineVidBench, for rigorously assessing Video-LLMs' performance at both the scene and fragment levels, offering a comprehensive evaluation of their capabilities.
We assessed multiple models and validated the effectiveness of SF$^2$T on them. Experimental results reveal that our approach improves their ability to capture and interpret spatiotemporal details.

\end{abstract}

%% file: 1_intro.tex
\section{Introduction}
\label{sec:intro}

\begin{figure}
    \centering
    \includegraphics[width=0.9\linewidth, page=1]{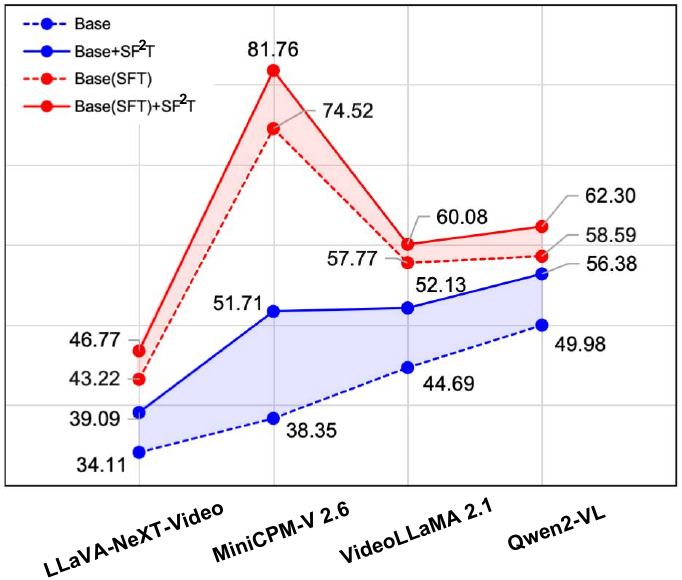}
    \caption{\textbf{Performance w/ and w/o SF$^2$T.} We evaluated four advanced Video-LLMs w/ and w/o SF$^2$T on our proposed FineVidBench with two baselines: (1) Base: performance without any fine-tuning (blue dashed), and (2) Base (SFT): performance with supervised fine-tuning (red dashed). After applying SF$^2$T, all models showed significant improvements (solid blue and red), underscoring its \textbf{broad} effectiveness.}
    \label{fig_intro}
\end{figure}

Large Language Models (LLMs) have showcased significant emergent capabilities, such as in-context learning~\cite{mann2020language}, instruction-following~\cite{raffel2020exploring}, and chain-of-thought reasoning~\cite{wei2022chain}, driven by expansive datasets and advanced model architectures. Extending these advancements, Video-LLMs through mechanisms like pooling or query aggregation across numerous visual tokens, have broadened the scope of LLMs to encompass video information processing~\cite{li2023videochat, zhang2023video,li2025llama}. This evolution markedly advances their potential for in-depth real-world comprehension, opening applications in intelligent surveillance, virtual reality, and autonomous driving, further enriching the landscape of video analytics and interpretation.

Various Video-LLMs, exemplified by GPT4-V, VideoLLaMA 2~\cite{cheng2024videollama}, MiniCPM-V~\cite{yao2024minicpm}, and Qwen2-VL~\cite{wang2024qwen2}, have been crafted by leading corporations and research institutions, demonstrating proficiency in capturing the overarching content of videos. When adapting to new videos and tasks, they predominantly rely on Supervised Fine-Tuning (SFT)~\cite{sun2019videobert} or Reinforcement Learning from Human Feedback (RLHF)~\cite{ziegler2019fine}, both of which are heavily contingent upon extensive manual annotation. This dependence poses several key problems: (1) it necessitates substantial human resources, particularly highly trained annotators; (2) the inherent complexity of video content and task demands frequently introduces inconsistencies and subjectivity, rendering the maintenance of high-quality annotations particularly arduous; and (3) subtle temporal variations across video frames are challenging to articulate with precision, often yielding generalized descriptions that constrain the Video-LLMs’ potential. Consequently, existing Video-LLMs struggle with fine-grained video understanding tasks, particularly in aspects such as visual dynamics (e.g., motion patterns, object interactions) and video details inquiries (e.g., positional changes, detail variations).

To address these challenges, we observe that fine-tuning Video-LLMs with self-supervised fragment tasks, by ``fragment'' we mean temporal frame level specifications of the video, could improve the model's sensitivity to spatiotemporal scene-level details (related to video contents). Driven by this, we introduce the \textbf{S}elf-\textbf{s}upervised \textbf{F}ragment \textbf{F}ine-Tuning (SF$^2$T), a effortless fine-tuning strategy for Video-LLMs that help to improve the fine-grained video understanding. SF$^2$T consists of five fragment-level tasks—Counting, Consistency Verification, Localization, Disorder Detection and Rearrangement—that automatically generate labels from various spatiotemporal perspectives. This approach maximizes the use of frame-level information while minimizing reliance on complex human instructions and annotations. 


Moreover, to evaluate the fine-grained visual dynamic perception of Video-LLMs and fully demonstrate the effectiveness of our SF$^2$T, we present the \textbf{FineVidBench}, a novel benchmark. FineVidBench comprises \textbf{910} videos and \textbf{22,718} question-answer pairs, with videos sourced from diverse public datasets, including Something-Something V2 (SSv2)~\cite{goyal2017something}, Moments in Time (MiT)~\cite{monfort2019moments}, etc. The question-answer pairs are auto-generated in single-choice format, incorporating distractors to increase testing difficulty.
We evaluated several notable Video-LLMs developed in recent years, and find they generally fail to understand the execution sequence of actions and struggling to grasp fine-grained spatiotemporal information. While after fine-tuning with SF$^2$T, the Video-LLMs better recognize spatiotemporal details, leading to a holistic and marked improvement in fine-grained understanding. 

%% file: 2_relatedwork.tex
\section{Related Work}
\label{sec:relatedwork}

\textbf{Video-LLMs Finetuning}
Video-LLMs are primarily fine-tuned by adjusting the parameters of small, trainable adapters for task adaptation, without changing the entire model, saving resources and enhancing efficiency. The connective adapter (e.g., MLP/Linear Layer~\cite{liu2024visual}, Q-former~\cite{li2023blip}) links the Video Embedder and LLM, aligning video embeddings with LLM input tokens, while insertive adapters (e.g., LoRA~\cite{hu2021lora}) are directly integrated into the LLM to modify its behavior. 
Most Video-LLMs combine both types of adapters and typically use multi-stage fine-tuning~\cite{zhang2023video,cheng2024videollama,li2023videochat,li2023llama,ren2024timechat}. First, the model learns to establish relationships between images, videos, and text using large-scale multimodal datasets~\cite{chen2024sharegpt4video,wang2023internvid,Alpher02,xu2023youku}. 
In the second stage, the model is fine-tuned with an curated instruction-following dataset~\cite{maaz2023video,li2023videochat,lyu2023macaw}. 
Besides, there are full fine-tuning, which updates all LLM parameters with a lower learning rate~\cite{shu2023audio,yang2023vid2seq}, and zero-shot models, which transforms the video task into a text task, typically relying on a powerful LLM~\cite{xu2024slowfast}.
However, annotating video data remains a labor-intensive and time-consuming task, particularly for long videos or those involving complex actions.

\noindent \textbf{Benchmarks on Video-LLMs}
Currently, many studies~\cite{fu2024video,chen2023autoeval,zhou2024mlvu} focus on evaluating the temporal perception capabilities of Video-LLMs. MVBench~\cite{li2024mvbench} designs 20 tasks from temporal and spatial perspectives, and Tempcompass~\cite{liu2024tempcompass} introduces 5 temporal aspects and 4 task formats. 
VNBench~\cite{zhao2024needle} decouples video content from the QA pairs by inserting irrelevant images or text ``needles'' into the original video.
Moment-10M~\cite{qian2024momentor} has constructed a large-scale dataset on temporal localization tasks. 
However, as illustrated in Table~\ref{table1}, these studies often focus on gathering diverse videos or evaluating the models' performance with long videos, while somewhat neglecting the models' ability to perform fine-grained perception of temporal details. 
To address this gap, FineVidBench breaks videos into multiple sets of frames and generates annotations from diverse spatiotemporal perspectives, introducing novel evaluation methods for fine-grained understanding.

\begin{table}[!htbp]\scriptsize
    \renewcommand{\arraystretch}{1.5}
    \setlength\tabcolsep{3pt}
    \centering
    \begin{tabular}{@{}l@{ }cccccc@{}}
        \toprule
        Benchmarks   & \makecell{Video\\num.} & \makecell{QA\\num.} & \makecell{Input\\Change} & \makecell{Temporal\\Diversity} & \makecell{Fine-Grained\\Evalution} & \makecell{Hierarchical\\Test} \\ 
        \hline
        Video-MME & 900    & 2700      & \textcolor{red}{\ding{55}}  & \textcolor{red}{\ding{55}}  &  \textcolor{red}{\ding{55}} & \textcolor{red}{\ding{55}}  \\
        TempCompass & 410    & 7540     & \textcolor{red}{\ding{55}}  & \textcolor{green}{\ding{51}} & \textcolor{green}{\ding{51}}  & \textcolor{red}{\ding{55}}  \\
        VNBench & -    & 1350      & \textcolor{red}{\ding{55}}  & \textcolor{green}{\ding{51}}  & \textcolor{green}{\ding{51}}  & \textcolor{red}{\ding{55}}  \\
        Moment-10M  & 64.9k  & 10.4M     & \textcolor{red}{\ding{55}}  & \textcolor{red}{\ding{55}}  & \textcolor{red}{\ding{55}}  & \textcolor{red}{\ding{55}}  \\
        AutoEval-Video  & 327    & 327   & \textcolor{red}{\ding{55}}  & \textcolor{red}{\ding{55}}  & \textcolor{red}{\ding{55}}  & \textcolor{red}{\ding{55}}  \\
        MVBench & 3641   & 4000      & \textcolor{red}{\ding{55}}  & \textcolor{red}{\ding{55}}  & \textcolor{green}{\ding{51}}  & \textcolor{red}{\ding{55}}  \\
        MLVU & 1334    & 2593      & \textcolor{red}{\ding{55}}  & \textcolor{red}{\ding{55}}  & \textcolor{red}{\ding{55}}  & \textcolor{red}{\ding{55}}  \\
        \hline
        \textbf{FineVidBench} & 910~  & 22,718~   & \textcolor{green}{\ding{51}} & \textcolor{green}{\ding{51}} & \textcolor{green}{\ding{51}} & \textcolor{green}{\ding{51}} \\
    \bottomrule
    \end{tabular}
    \caption{Comparison with related benchmarks. Our approach offers significant advantages in input formats, evaluation methods, granularity, and temporal diversity.}
    \label{table1}
\end{table}

%% file: 4_benchmark.tex
\section{FineVidBench Benchmark}
\label{sec:benchmark}

It is broadly recognized that Video-LLMs struggle with fine-grained video understanding tasks, yet no comprehensive benchmarks exist to thoroughly investigate this issue. To address this gap, we introduce FineVidBench, a multidimensional, fine-grained evaluation framework specifically designed to assess and improve the overall capabilities of Video-LLMs.

\subsection{Construction}



\noindent \textbf{Data collection} We selected videos from various public datasets, including SS-v2~\cite{goyal2017something}, MiT~\cite{monfort2019moments}, and Ego4D~\cite{grauman2022ego4d}, with a particular emphasis on temporally-sensitive content, to focus the model on the entire video sequence rather than individual frames.

\begin{figure}
    \centering
    \includegraphics[width=0.9\linewidth, page=2]{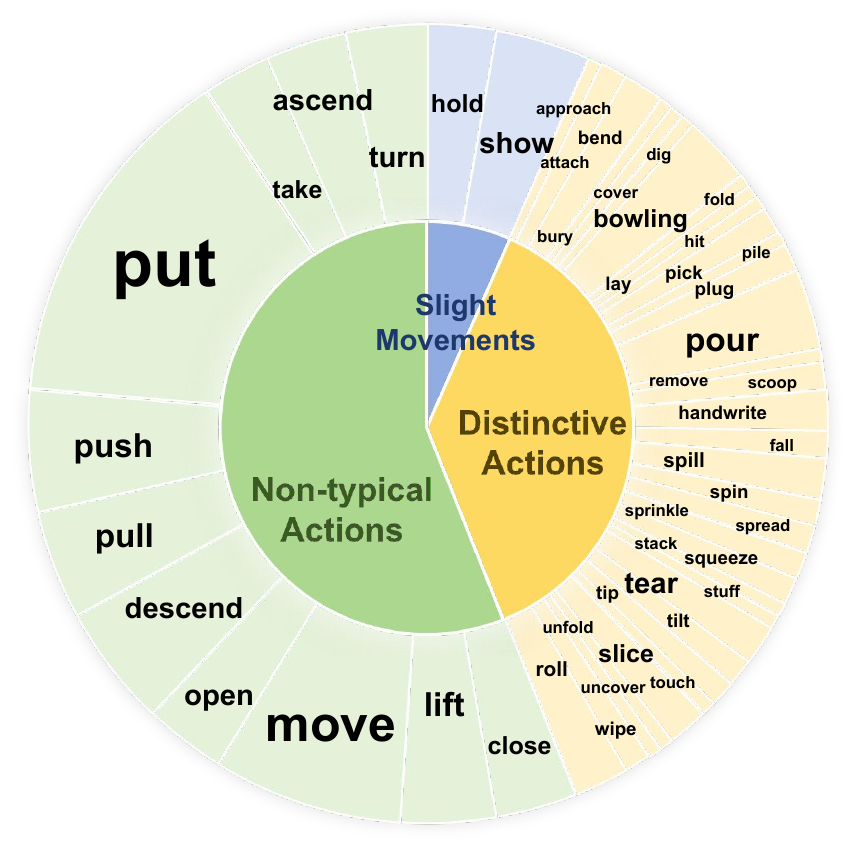}
    \caption{We show the action semantics and their respective proportions in FineVidBench. Distinctive Action: easily recognizable actions. Non-typical Action: flexible actions with no clear characteristics, like ``put'' and ``move.'' Slight Movement: subtle actions, such as ``hold'' and ``show,'' difficult to detect with the naked eye.}
    \label{fig_bench_actiontype}
\end{figure}

\noindent \textbf{Action categorization} As shown in Figure~\ref{fig_bench_actiontype}, we compiled 52 actions, categorizing them into 3 types based on intra-class variance. The distribution varies significantly:
``Distinctive Actions'' (39\%) are easily recognizable, encompassing a total of 36 actions.
``Non-typical Actions'' (57\%) refer to flexible actions with no clear defining characteristics, spanning 14 types. The broad diversity and complexity in this category require more extensive video coverage to adequately capture the range of expressions and variations.
``Slight Movements'' (4\%) represent subtle actions, such as ``hold'' and ``show'', which are difficult to detect with the naked eye and constitute a small proportion.


\noindent \textbf{Data augmentation} The original videos were augmented using frame interpolation and skipping techniques for speed transformation, along with a motion-salient area sampling algorithm to capture dynamic motion. This process generated speed-varied versions and multiple sets of keyframes for each video.

\noindent \textbf{Statistics} With our augmentation strategy, FineVidBench includes 910 videos, 1,820 speed-variant videos, and 2,670 sets of keyframes enriched with dynamic visual information. Building on this, we generated 22,718 QA pairs from the video content through a combination of automated processes and manual review. The quality assurance process involved rigorous cross-verification, where reviewers checked each QA pair for accuracy and contextual relevance, making corrections to ensure high quality.


\begin{figure*}[!ht]
    \centering
    \includegraphics[width=0.95\linewidth, page=3]{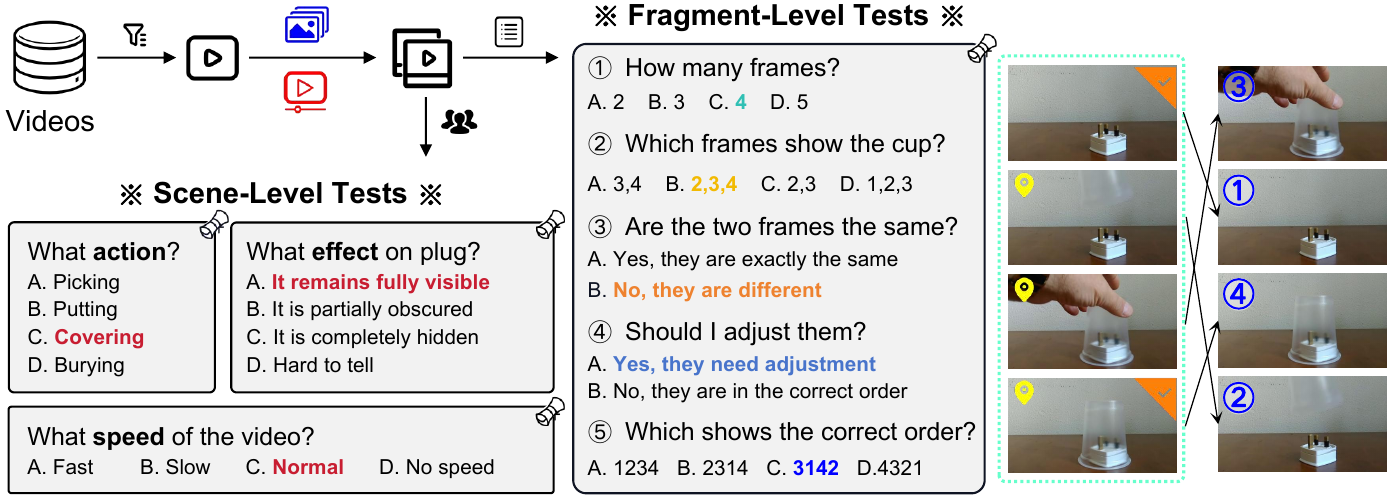}
    \caption{FineVidBench evaluates videos \textbf{augmented} with speed variations and fragments. \textbf{Scene-level tests} include the following: Action: Tests recognition accuracy amidst distractors like ``Visual Synonyms''. Effect: Assesses the model’s ability to identify pre- and post-action changes. Speed: Measures the model’s sensitivity to changes in video speed. \textbf{Fragment-level tests}, employing a step-by-step inquiry framework, focus on challenges such as Frame Count, Meaning of Order, Frame Comparison, Adjust-or-Not and Rearrangement.}
    \label{fig_bench}
\end{figure*}

\subsection{Benchmarking Dimensions}

As shown in Figure~\ref{fig_bench}, FineVidBench encompasses both scene-level and fragment-level evaluations. The \textbf{scene-level} evaluation assesses both original and speed-adjusted videos across three dimensions: (1) \textbf{Action}, which evaluates the model’s holistic understanding of video content. To increase difficulty, ``Visual Synonyms'' are added as distractors, requiring VideoLLM to distinguish visually similar actions with subtle differences, a challenge common in real-world scenarios. (2) \textbf{Effect}, which focuses on the model’s comprehension of the visual changes resulting from actions. This understanding is essential for revealing object properties and interpreting complex dynamic scenes, and could significantly enhance the reasoning capabilities of Video-LLMs and LLM-aided agents. (3) \textbf{Speed}, which tests the model’s sensitivity to changes in video speed and its capability to maintain consistent understanding across varying speeds, with slow motion revealing hidden details and fast motion obscuring them. This capability is crucial for optimizing the model’s performance across diverse scenarios.

For \textbf{fragment-level} evaluation, We’ve designed a structured evaluation format for video dynamic keyframes, employing a \textbf{step-by-step} inquiry framework: 
(1) Frame Count: Models are queried on the number of frames in sequences using dynamically refined keyframes to assess counting accuracy.
(2) Meaning of Order: Understanding of sequence order is tested by asking about the first or last frames the targets appear in, or the frames they are present. e.g., ``At which frame does the target object first appear?''. 
(3) Frame Comparison: Two frames are randomly selected from the sequence for visual comparison, with differences varying in size but generally staying within human visual comfort limits.
(4) Adjust-or-Not and Rearrangement: These two tasks involve a shuffled sequence of keyframes, and the model is asked to determine whether the order needs adjustment and, if so, how to correct it. They evaluate the model's ability to understand and restore the video's temporal sequence.


\subsection{Benchmark Results}

We evaluated \textbf{six} of the most advanced open-source models: LLaVA-NeXT-Video\cite{li2024llava}, MiniCPM-V 2.6\cite{yao2024minicpm}, VideoLLaMA 2.1\cite{cheng2024videollama}, Qwen2-VL\cite{wang2024qwen2}, ShareGPT4Video~\cite{chen2024sharegpt4video} and Video-CCAM~\cite{tencent2024video-ccam}, each employing different architectures and training strategies. Table~\ref{tab_bench_all} summarizes the results across the eight tasks. We discuss the results from scene-level and fragment-level. 


\subsubsection*{$\bullet$ Scene-level Results and Analysis}

\noindent \textit{Action} The scores for this task varied significantly, with models trained in relevant video data—such as Video-CCAM, Qwen2-VL, and VideoLLaMA 2.1—achieving notably higher performance. However, as shown on the left side of  Table~\ref{tab_bench_Ac&FC}, interference from ``Visual Synonyms'' prevented these models from achieving their full potential, resulting in declines of varying degrees and indicating difficulties in distinguishing visually similar actions.

\noindent \textit{Effect} All models exhibited average performance on this task, indicating a superficial understanding of aspects such as object attributes, object relationships, and action properties. 
This task tests the model's ability to grasp how actions affect objects, focusing on causal relationships and temporal reasoning—particularly for actions like ``push'' and ``pull'', which share similar execution flows. The model must distinguish them based on dynamic effects, such as changes in direction and speed, but most models perform moderately in this regard.

\begin{figure}
    \centering
    \includegraphics[width=0.85\linewidth, page=4]{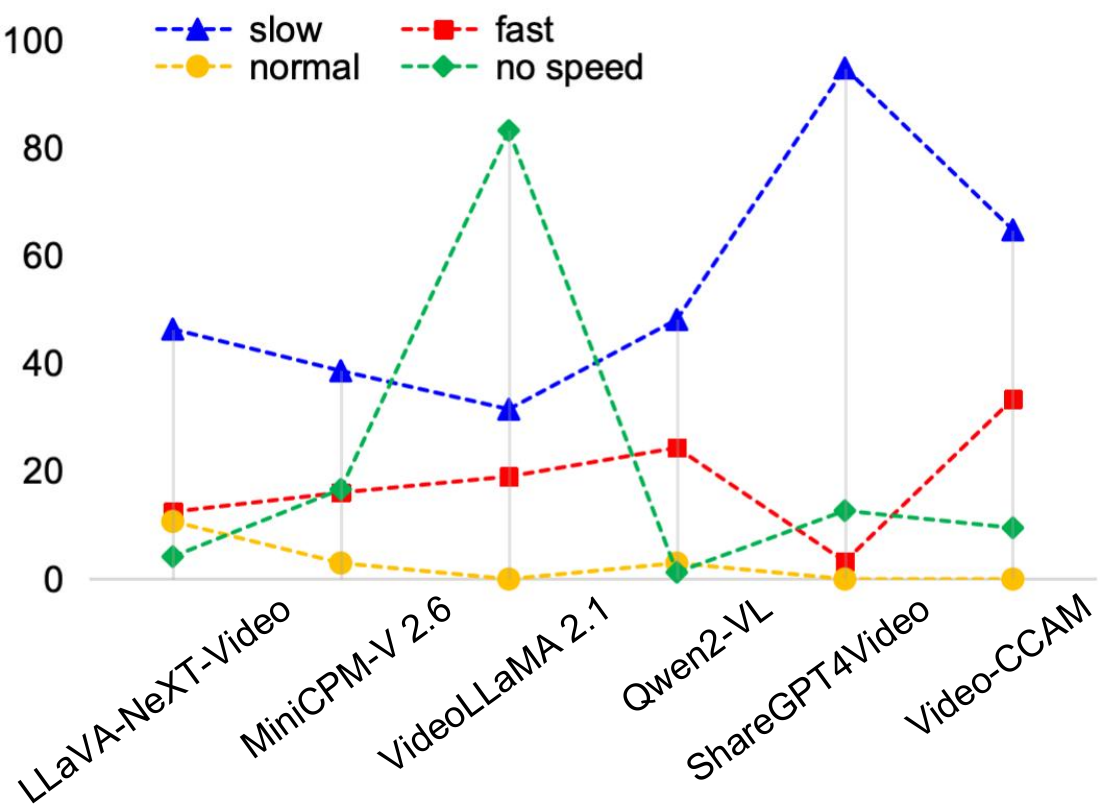}
    \caption{Accuracy across different video speeds. All models are more sensitive to slow-speed videos and struggle to understand ``normal speed'' and ``no speed'', except for VideoLLaMA 2.1.}
    \label{fig_bench_speed}
\end{figure}

\noindent \textit{Speed} The results show that all models are insensitive to speed variations, likely because they were not adequately exposed to speed changes during training. Figure~\ref{fig_bench_speed} shows that models are more sensitive to slow motion than fast playback, and struggled with identifying ``normal speed'' and ``no speed'', except for VideoLLaMA 2.1. This may be due to the loss of coherence in fast-moving video content, while slow-motion videos highlight more distinct details, aiding the model in making accurate judgments.

\begin{table}
    \renewcommand{\arraystretch}{1.3}
    \setlength
    \tabcolsep{3pt}
    \centering
    \arrayrulecolor{black}
    \centering
    \small
    \definecolor{lightgrey}{rgb}{0.95, 0.95, 0.95}
    \setlength{\tabcolsep}{1mm}{
    \begin{tabular}{l|c@{ }>{\columncolor{lightgrey}}c@{ }|c@{ }>{\columncolor{lightgrey}}c@{ }>{\columncolor{lightgrey}}c@{}>{\columncolor{lightgrey}}c@{ }}
        \toprule
        \multirow{2}{*}{Video-LLMs}  & \multicolumn{2}{c|}{Action} &\multicolumn{4}{c}{Frame Number}  \\
        & w/o VS~ & w/ VS~ & Avg. & 3 & 4 & 5 \\
        \hline
        LLaVA-NeXT-Video  & 37.31~  & 35.04~  & 19.37~ & 20.33~ & 19.77~ & 17.98~ \\
        \cline{1-1}
        MiniCPM-V 2.6     & 43.37~  & 40.15~  & 90.32~ & 93.82~ & 90.66~ & 86.44~\\
        \cline{1-1}
        Video-LLaMA 2.1   & 63.26~  & 53.98~  & 30.17~ & 42.86~ & 39.89~ & 7.45~\\
        \cline{1-1}
        Qwen2-VL       & 68.18~  & 56.62~  & 96.65~ & 97.25~ & 96.63~ & 96.05~\\
        \cline{1-1}
        ShareGPT4Video & 46.90~ & 30.84~ & 26.33~ & 60.99~ & 16.78~ & 0.00~ \\
        \cline{1-1}
        Video-CCAM & 73.10~ & 60.23~ & 23.45~ & 14.18~ & 8.96~ & 47.61~\\
        \bottomrule
    \end{tabular}
    \caption{\textbf{Left}: Accuracy of the Action task with or without ``Visual Synonyms''. It is obvious that the ``Visual Synonyms'' have significantly impacted the model's judgment. \textbf{Right}: Accuracy of the counting task across different frame counts. Except for Video-CCAM, all other models exhibited a decline in performance as the number of frames increased.}
    \label{tab_bench_Ac&FC}}
\end{table}

\begin{table*}[!htbp]
    \renewcommand{\arraystretch}{1.5}
    \setlength
    \tabcolsep{2.5pt}
    \centering
    \arrayrulecolor{black}
    \begin{tabular}{lc|ccc|ccccc|c|c|c} 
        \toprule
        \multirow{2}{*}{Video-LLMs}  & \multirow{2}{*}{Params.} & \multicolumn{3}{c|}{Scene-Level}       & \multicolumn{5}{c|}{Fragment-Level}       & \multirow{2}{*}{S-Avg.} & \multirow{2}{*}{FG-Avg.} & \multicolumn{1}{c}{\multirow{2}{*}{A-Avg.}}  \\ 
        \arrayrulecolor{black}\cline{3-10}
    &    & Action   & Effect   & Speed    & FCnt     & MoO      & FCmp     & AoN      & Rearr    &   &    & \multicolumn{1}{c}{}   \\ 
        \arrayrulecolor{black}\hline
        \rowcolor[rgb]{0.97,0.97,0.97} (Random) & -  & 25.00~   & 25.00~   & 25.00~   & 25.00~   & 25.00~   & 33.33~   & 33.33~   & 25.00~   & 25.00~   & 28.33~    & 27.08~   \\ 
        \hline
        LLaVA-NeXT-Video     & 7B  & 37.31~  & 42.67~  & 22.35~    & 19.37~  &24.02~  &53.75~  &75.45~  &20.67~      &34.11~  &38.65~  &36.95~   \\ 
        \arrayrulecolor{black}\cline{1-2}
        MiniCPM-V 2.6  & 8B  &43.37~  & 52.56~  &19.13~    &\uline{90.32}~  & \uline{56.42}~  & 75.66~  &76.49~  &18.09~      &38.35~  &\uline{63.40}~  &\uline{54.01}~   \\ 
        \cline{1-2}
        Video-LLaMA 2.1  & 7B  & 63.26~  &50.92~  & 19.89~    &30.17~ & 42.27~ 	&\uline{76.01}~  & 89.92~  &\textcolor{red}{26.87~}      & 44.69~  &53.05~  &49.91~  \\ 
        \cline{1-2}
        Qwen2-VL  & 7B  & \uline{68.18}~  & \textcolor{red}{57.14}~   & 24.62~   & \textcolor{red}{96.65}~   & 33.33~   & 74.53~   & \textcolor{red}{90.70}~   & 22.48~   & \uline{49.98}~   & \textcolor{red}{63.54}~    & \textcolor{red}{58.45}~   \\
        \cline{1-2}
        ShareGPT4Video & 8B & 46.90~   & 43.88~   & \textcolor{red}{31.76}~ & 26.33~ & \textcolor{red}{61.05}~ & \textcolor{red}{88.44}~ & 84.80~   & \uline{23.36}~   & 40.85~   & 57.11~  & 50.82~ \\
        \cline{1-2}
        Video-CCAM & 9B & \textcolor{red}{73.10}~   & \uline{55.90}~ & \uline{31.65}~   & 23.45~   & 45.66~   & 64.95~   & \uline{90.27}~ & 22.72~   & \textcolor{red}{53.55}~ & 48.47~    & 50.96~ \\
        \bottomrule
    \end{tabular}
    \caption{The overall performances of notable Video-LLMs on FineVidBench. FCnt: Frame Count. MoO: Meaning of Order. FCmp: Frame Comparison. AoN: Adjust or Not. Rearr: Rearrangement. S-Avg.: the average performance of scene-level tasks; FG-Avg.: the average performance of fragment-level tasks. A-Avg.: the average performance of all tasks.}
    \label{tab_bench_all}
\end{table*}

\subsubsection*{$\bullet$ Fragment-level Results and Analysis}

(1) Frame-count accuracy varied significantly across models, with the lower-performing models likely lacking targeted training. The trend shown in the right side of Table~\ref{tab_bench_Ac&FC}, where accuracy decreases as frame count increases, highlights the models’ insufficient temporal reasoning on longer sequences. 
(2) ShareGPT4Video and MiniCPM-V 2.6 showed better comprehension in the Meaning-of-Order task, while other models lagged, suggesting a lack of explicit focus on ``order''. 
(3) Most models excelled in frame comparison due to image-text alignment training. ShareGPT4Video achieved the best performance, owing to its Differential Sliding-Window Captioning (DiffSW) strategy, which emphasizes capturing the changes between frames when generating video descriptions. This also improved its Meaning-of-Order performance.
(4) In the sorting task, models generally succeeded in the ``Adjust or Not'' response but performed poorly in the more complex ``Rearrangement'' task, indicating they can detect, but not correct, sequence errors.

%% file: 3_method.tex
\section{Self-supervised Fragment Finetuning}
\label{sec:method}

\begin{figure}
    \centering
    \includegraphics[width=0.98\linewidth]{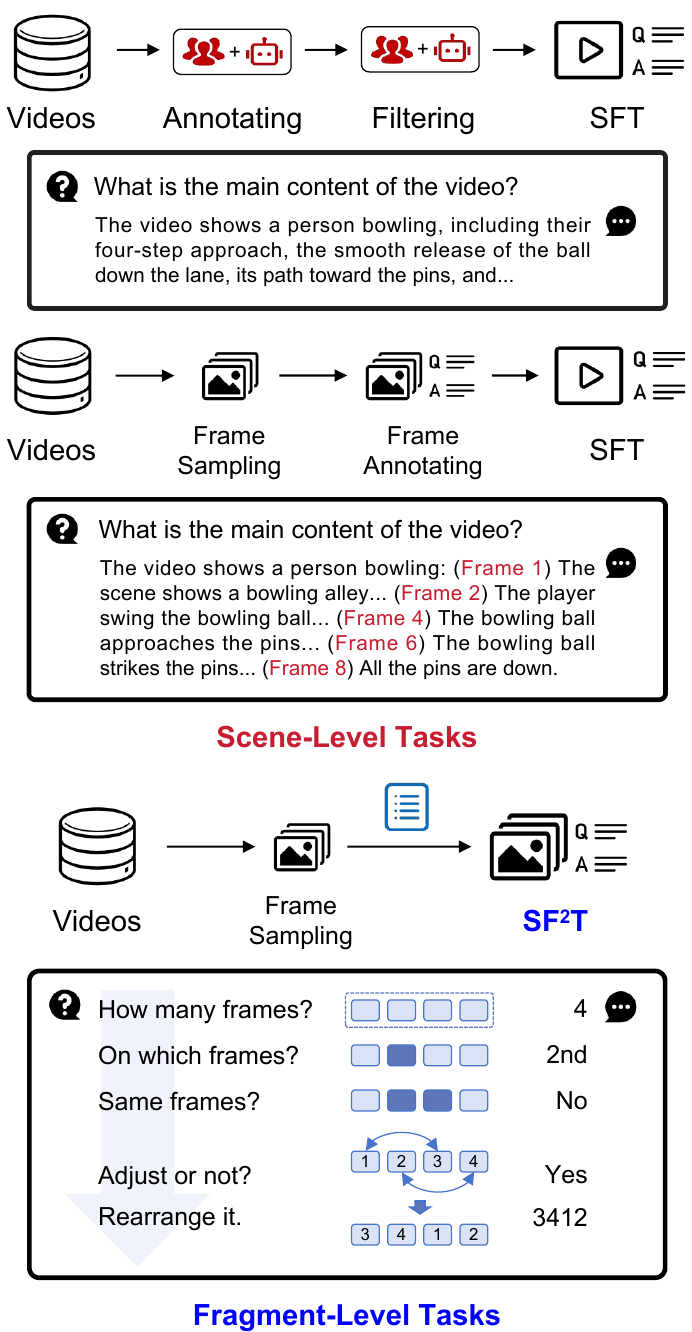}
    \caption{Comparison between SF$^2$T and SFT. SFT depends on \textbf{manual and model-driven design} to generate QA pairs for scene-level video understanding, SF$^2$T, in contrast, automatically constructs training data based on \textbf{pre-defined rules} that cover various temporal and spatial aspects of the video. SF$^2$T enables the model to focus on a fine-grained content analysis, and offering insights that supervised labels \textbf{cannot} achieve.}
    \label{fig_method}
\end{figure}

 
The above benchmark results show the existing Video-LLMs generally fail to tackle fine-grained video understanding tasks. 
%
%
Videos often contain subtle, complex changes that natural language alone fails to fully capture. The core component of Video-LLMs, LLMs, as generalized pattern recognizers, offers a promising solution. LLMs have the potential to detect and interpret intricate spatiotemporal dynamics that were previously difficult to represent. 
Given that these changes cannot be directly annotated, using self-supervised learning naturally becomes the solution, bypassing the bottleneck of manual annotation and significantly reducing labeling costs. 
Given these factors, we propose the SF$^2$T to fine-tune Video-LLMs. While we do not expect SF$^2$T to replace the supervised fine-tuning, instead it's \textbf{an effortless complementary to SFT}. Comparing SF$^2$T with SFT, they primarily differ in data construction and content focus level, with each method aligned with distinct training objectives as shown in Figure~\ref{fig_method}.

\subsection{SFT Tasks}

We first review the common SFT tasks to set a baseline for comparing our SF$^2$T.

\noindent\textbf{General QA on Video Content}
This method focuses on understanding the main events and context of a video by directly asking questions about its content. While effective for grasping the video's key moments, it lacks finer spatiotemporal details and requires significant human effort to create standardized but constrained answers.

\noindent\textbf{Frame Description Integration}
This method typically samples video frames evenly, generates detailed descriptions for each, and integrates them into a cohesive but lengthy summary. While it enhances the model's understanding of continuity and micro-dynamics, it often proves incapable of capturing complex or subtle details that are beyond natural language's scope. Moreover, although frame descriptions can be generated using powerful multi-model LLMs like GPT-4o, significant human effort is still required to review the quality of the generated responses.


\subsection{Fragment-level Tasks of SF$^2$T}
SFT tasks require manual annotations, and even automation annotation is labor-intensive and error-prone. To address, we introduce SF$^2$T which generates accurate fragment-level labels accurately. SF$^2$T comprises five tasks—Counting, Consistency Verification, Localization, Disorder Detection and Rearrangement—designed to train the model to rearrange a set of out-of-order frames into their original sequence. This is a robust indicator of a modal’s mastery over the visual dynamics of an action, requiring the model to detect subtle frame changes and understand the overall coherence and temporal trends. Mastery of these tasks enables the model to recognize frames and their temporal relationships, enhancing its ability to predict and reconstruct action sequences and improving performance on more complex video tasks.
%
%
Our method first extracts multiple sets of dynamic keyframes from each video. These fragments capture the key dynamic information from multiple temporal perspectives, offering a more efficient representation of redundant video data. It then applies pseudo-labeling, distinguishing it from traditional video-level labeling. By designing proxy tasks that leverage intrinsic information rather than predefined prior knowledge, it smartly circumvents the annotation bottleneck, enabling a deeper temporal understanding and offering insights that traditional video-level labeling cannot achieve.

\noindent \textbf{Counting}
%
We input N frames into the Video-LLM and ask it to count them. Although this task seems straightforward, it proves challenging for current Video-LLMs, particularly as the number of frames increases, revealing a decline in accuracy. 
The model's inability to perform basic quantitative tasks points to a broader limitations in understanding the overall sequence integrity.

\noindent \textbf{Consistency Verification}
Video-LLMs are tasked with identifying two frames sampled from the same video, which may show subtle differences. This task sharpens the model's sensitivity to visual details by encouraging a thorough analysis and comparison of the images, countering its tendency to focus on primary subjects while neglecting the background and other subtle features.

\noindent \textbf{Localization}
Video-LLMs must accurately locate a specified target (from video metadata) within a sequence of frames, identifying the frames in which it appears, disappears, or persists. This naturally human ability is a significant challenge for these models, as they often struggle to perceive sequential relationships between frames and face additional obstacles, such as occlusion, interference from similar objects, lighting variations, and memory limitations.

\noindent \textbf{Disorder Detection and Rearrangement}
Video-LLMs must determine whether and how to adjust the order of a given frame sequence. When frames are randomized, the loss of spatiotemporal coherence and logical continuity makes it exceptionally challenging to reconstruct their original sequence, especially as interactions within frames become more complex~\cite{misra2016shuffle}. This task is evaluated in two ways: the yes/no task tests the model's sensitivity to temporal consistency, while the sorting task, which leverages capabilities from the other four tasks, requires advanced reasoning and adjustments.



%% file: 5_experiment.tex
\section{Experiments}
\label{sec:experi}

\begin{table*}[!htbp]
    \renewcommand{\arraystretch}{1.5}
    \setlength
    \tabcolsep{2.5pt}
    \centering
    \arrayrulecolor{black}
    \definecolor{lightgrey}{rgb}{0.95, 0.95, 0.95}
    \begin{tabular}{l|ccc|ccc|ccc|ccc} 
        \toprule
        \multirow{2}{*}{Methods} & \multicolumn{3}{c|}{LLaVA-NEXT-Video} & \multicolumn{3}{c|}{MiniCPM-V 2.6} & \multicolumn{3}{c|}{VideoLLaMA 2.1} & \multicolumn{3}{c}{Qwen2-VL}   \\
        \arrayrulecolor{black}\cline{2-13}
     &Action  & Effect & Speed &Action  & Effect & Speed  &Action  & Effect & Speed  &Action  & Effect & Speed   \\ 
        \hline
        Base & 37.31~ & 42.67~  &22.35~    & 43.37~ & 52.56~ & 19.13~   & 63.26~ &50.92~  &19.89~   & 68.18~ & 57.14~ & 24.62~ \\
        Base+SF$^2$T & \textbf{48.67}~ & \textbf{43.77}~  &\textbf{24.83}~  & \textbf{65.91}~ & \textbf{60.62}~ & \textbf{28.60}~   & \textbf{67.42}~ & \textbf{57.33}~  & \textbf{31.63}~  & \textbf{73.86}~ & \textbf{63.37}~  & \textbf{31.92}~  \\
        \hline\hline
        Base(SFT)  & 62.69~ & 44.63~   & 22.35~  & 77.65~ & 75.09~  & 70.83~  & 77.65~ & 65.94~  & 29.73~ & 78.60~ & 66.30~ & 30.87~  \\
        Base(SFT)+SF$^2$T & \textbf{63.07}~ & \textbf{45.24}~   & \textbf{32.01}~  & \textbf{81.63}~ & \textbf{76.92}~  & \textbf{86.74}~  & \textbf{79.73}~ & \textbf{68.68}~  & \textbf{31.82}~ & \textbf{81.25}~ & \textbf{73.26}~ &\textbf{32.38}~ \\ 
        \bottomrule
    \end{tabular}
    \caption{\textbf{Performance on FineVidBench.} We tested on two baselines: (1) Base: Results without any fine-tuning. (2) Base(SFT): Results after fine-tuning in supervised way. 
    After SF$^2$T, all models improved in all three tasks, highlighting its broad effectiveness and the value of fragment-level tasks in enhancing scene-level comprehension. Notably, SF$^2$T outperformed SFT in the Speed task (except MiniCPM-V 2.6), highlighting the key role of fine-grained temporal understanding in distinguishing video speeds.}
    \label{tab_exp_all}
\end{table*}

\begin{table*}[!htbp]
    \renewcommand{\arraystretch}{1.5}
    \setlength
    \tabcolsep{2.5pt}
    \centering
    \arrayrulecolor{black}
    
    \begin{minipage}{.48\textwidth}
    \centering
    \renewcommand{\arraystretch}{1.2}
    \setlength
    \tabcolsep{2.5pt}
    \small
    \begin{tabular}{l|c|c|c|c} 
        \toprule
        \multirow{1}{*}{Methods} & \footnotesize\makecell{LLaVA-NeXT\\-Video~} & \footnotesize\makecell{MiniCPM-V\\2.6~} & \footnotesize\makecell{VideoLLaMA\\2.1~} & \footnotesize\makecell{Qwen2\\-VL~} \\
        \hline
        \multicolumn{5}{l}{\textit{\footnotesize\textcolor{gray}{MVBench}}} \\
        \hline
        Base &36.84~  &40.23~  &54.18~  &55.97~  \\
        Base+SF$^2$T &\textbf{42.92}~  &\textbf{56.02}~  &\textbf{57.97}~  &\textbf{63.76}~  \\
        \hline
        \multicolumn{5}{l}{\textit{\footnotesize\textcolor{gray}{Video-MME(no subtitle)}}} \\
        \hline
        Base &29.76~  &43.17~  &49.02~  &43.77~  \\
        Base+SF$^2$T &\textbf{34.84}~  &\textbf{53.19}~  &\textbf{51.88}~  &\textbf{53.60}~  \\
        \hline
        \multicolumn{5}{l}{\textit{\footnotesize\textcolor{gray}{MLVU}}} \\
        \hline
        Base &36.32~  &41.58~  &52.32~  &42.81~  \\
        Base+SF$^2$T &\textbf{41.91}~  &\textbf{55.32}~  &\textbf{56.11}~  &\textbf{54.67}~  \\
        \bottomrule
    \end{tabular}
    \captionsetup{width=1\textwidth}
    \caption{\textbf{Performance on public benchmarks.} SF$^2$T consistently enhances performance across all three benchmarks, reaffirming its effectiveness as a spatiotemporal enhancer.}
    \label{tab_mvbench}
    \end{minipage}%
    \hfill 
    \begin{minipage}{.48\textwidth}
    \small
    \centering
        \begin{tabular}{c|c|c|c|c} 
        \toprule
        Methods & random & uniform & keyframe & motion-salient~  \\ 
        \hline
        SF$^2$T & 70.31~ & 71.67~ & 72.11~  & 73.86~ \\
        \bottomrule
    \end{tabular}
    \captionsetup{width=1\textwidth}
    \caption{Impact of sampling. As shown, motion-salient area sampling outperforms others by better capturing motion fluidity and temporal details, while the other methods fail to fully utilize their potential, leading to suboptimal performance.}
    \label{tab_exp_samp}
    \vspace{0.2cm} 
    \begin{tabular}{c|c|c|c} 
        \toprule
        Methods & long & short & random  \\ 
        \hline
        SF$^2$T & 69.38~ & 71.40~  & 73.86~ \\
        \bottomrule
    \end{tabular}
    \captionsetup{width=1\textwidth}
    \caption{Impact of temporal span. Both long- and short-range temporal modeling reduced SF$^2$T's performance, emphasizing the importance of multi-scale temporal modeling.}
    \label{tab_exp_longshort}
    \end{minipage}
\end{table*}

In this section, we fine-tuned \textbf{four} of the most advanced open-source Video-LLMs using the SF$^2$T method to evaluate its effectiveness, alongside ablation studies and interpretability analyses to explore the underlying mechanisms.

\subsection{Implementation Details}

To ensure fairness, experiments were conducted on LoRA-compatible models, including LLaVA-NeXT-Video\cite{li2024llava}, MiniCPM-V 2.6\cite{yao2024minicpm}, VideoLLaMA 2.1\cite{cheng2024videollama} and Qwen2-VL\cite{wang2024qwen2}, using their default or recommended settings, with all models trained for one epoch. All experiments were performed under identical hardware conditions, utilizing NVIDIA A100 40GB GPU for computation. It should be emphasized that our goal is to validate the effectiveness of SF$^2$T, not to optimize models for maximum performance.

We randomly sampled videos from SSv2 and MiT for training, ensuring no overlap with the FineVidBench dataset. 
MGSampler~\cite{zhi2021mgsampler} was used to extract N sets of M-frame sequences from each video, capturing dynamic changes while preserving overall characteristics. 
M is chosen based on the video's characteristics to capture content flow, while N is determined by content complexity, with more complex content requiring a larger N to cover more temporal perspectives. 
In this study, we set N = 3 and M between 3 and 5, though these values may vary for other datasets. 
We then generated QA pairs for each frame sequence based on the five tasks defined in SF$^2$T for training. 
Evaluations were performed on FineVidBench's scene-level tasks, including Action, Effect, and Speed. 
To compare with traditional SFT, we also generated and manually reviewed QA pairs for these videos in a supervised setting.

\subsection{Comparisons}

Table~\ref{tab_exp_all} summarizes the results of the scene-level tasks. After SF$^2$T training, all models showed significant improvement, emphasizing that fragment-level tasks can notably enhance scene-level comprehension. Integrating SF$^2$T with SFT is also leads to performance gains, demonstrating that fragment-level training positively impacts SFT and enhances its effectiveness. 
Surprisingly, in the Speed task, many base models outperformed SFT after applying SF$^2$T, highlighting the importance of fine-grained temporal understanding in distinguishing video speeds. This improvement likely stems from SF$^2$T's ability to enhance the model's sensitivity to temporal cues, such as the loss or enhancement of information during acceleration or deceleration, as well as content coherence—all crucial for speed judgment.
As expected, SF$^2$T currently lags behind SFT, since its training objective is not fully aligned with scene-level tasks. However, we do not expect SF$^2$T to replace supervised fine-tuning; rather, our experiments suggest that it can serve as an effortless and effective complement to SFT.


In addition to FineVidBench, we evaluated SF$^2$T on three public video understanding benchmarks (Table~\ref{tab_mvbench}). The results demonstrate consistent improvements across various video tasks, validating SF$^2$T as an effective spatiotemporal enhancer for a wide range of video understanding tasks. All models were tested with an 8-frame input. 




\subsection{Ablation and Interpretability Analyses}

We evaluated the impact of frame sampling strategies on SF$^2$T, as each method provides a unique ``temporal information perspective'' that influencing video understanding performance. As shown in Table~\ref{tab_exp_samp}, we assessed four strategies on Qwen2-VL in the Action task: random, uniform interval, keyframe, and motion-salient area sampling~\cite{zhi2021mgsampler}. Motion-salient area sampling performed best, likely due to its ability to capture continuous motion dynamics, thereby enhancing the model’s understanding of action fluidity and temporal detail. In comparison, the other methods had limitations: keyframe sampling misses intermediate action phases, fixed-interval sampling may overlook critical moments, and random sampling lacks temporal consistency. Notably, different datasets may favor different strategies. For example, some datasets may perform better with uniform interval sampling, or their motion features may align better with the model’s specific capabilities.

We examined the effects of long- and short-range temporal modeling on SF$^2$T. In the Consistency Verification task, we constrained the random selection of frame pairs to adjacent frames for local continuity or non-adjacent frames to capture long-range dependencies.  As shown in Table~\ref{tab_exp_longshort}, both settings decreased SF$^2$T's performance on the Action task of Qwen2-VL, indicating that an overemphasis on either long- or short-range information leads to temporal imbalance and incomplete dynamics. This underscores the importance of combining both approaches to leverage their broader temporal span and frame variations for a more comprehensive feature representation.

\begin{figure}
    \centering
    \includegraphics[width=0.9\linewidth, page=2]{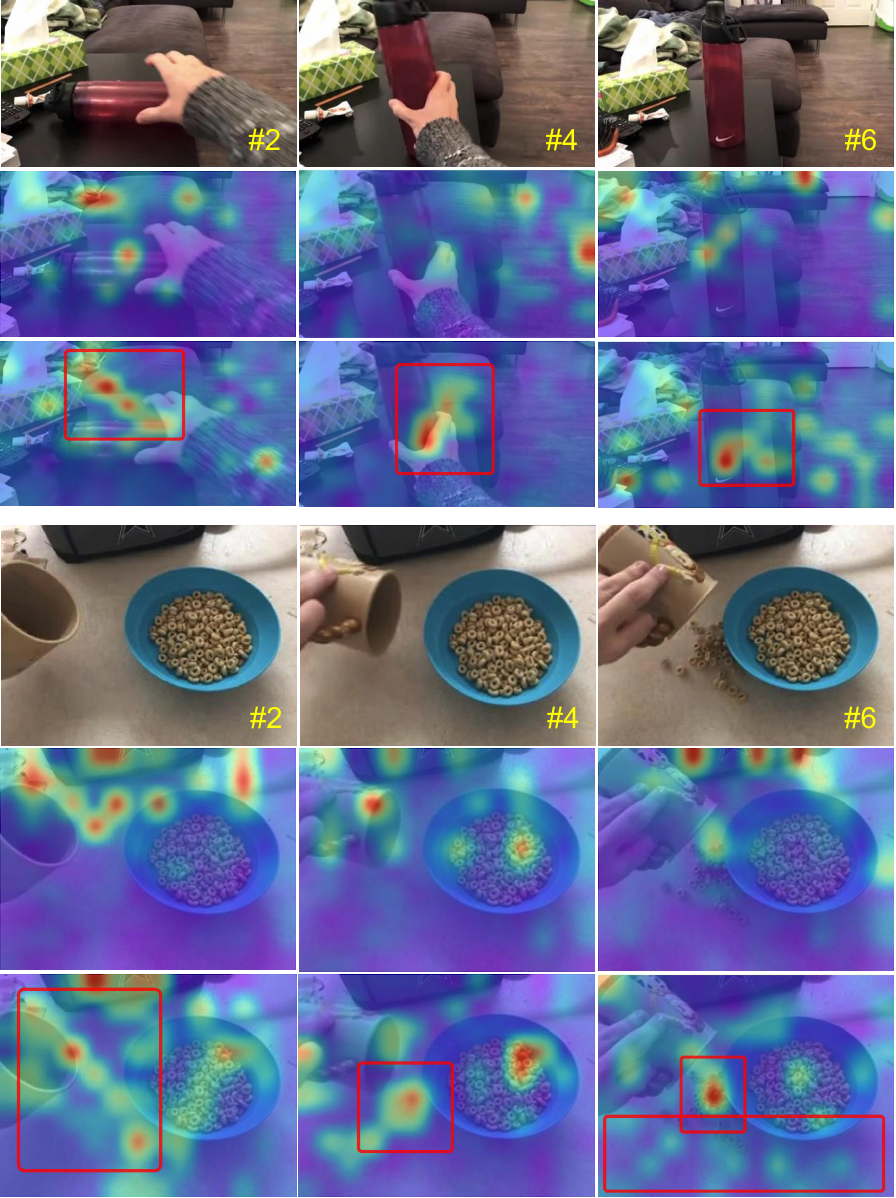}
    \caption{Two exemplary visualizations of the attention map on Qwen2-VL. For each example: top - Original frames; middle - Base (SFT); bottom - SF$^2$T applied. As shown by the red boxes, after applying SF$^2$T, the model better focuses on action execution areas and interacting objects. The SF$^2$T fine-tuned model has the ability to \textbf{predict the direction of motion}, as seen in the trajectories of the red bottle and Cheerios.}
    \label{fig_exp_attnmap}
\end{figure}

We analyzed the attention map of Qwen2-VL on the Action task, particularly in cases where the model's predictions were corrected after SF$^2$T. As shown in Figure~\ref{fig_exp_attnmap}, we found that SF$^2$T enhances the model's ability to capture fine-grained spatial changes and temporal dynamics.
(1) \textbf{Spatial Aspects.} After SF$^2$T, the model shows increased attention to action execution areas, particularly the hands and objects they interact with. It shows better sensitivity to small targets, likely due to the Consistency Verification task, which enhances spatial perception by refining sensitivity to subtle image differences.
(2) \textbf{Temporal Aspects.} After SF$^2$T, we observed that the model can predict object movement trajectories in certain actions, indicating an advanced level of temporal understanding. This ability likely stems from the sorting task, which strengthens the model’s comprehension of action flows and movement patterns.


%% file: 6_conclusion.tex
\section{Conclusion}
\label{sec:conc}
In this work, we propose SF$^2$T to overcome the limitations of Video-LLMs in fine-grained video understanding. SF$^2$T is an innovative fine-tuning method that eliminates the need for labor-intensive annotations and effectively bypasses the constraints of natural language descriptions. Additionally, we introduce FineVidBench, a benchmark for evaluating Video-LLMs at both scene and fragment levels. In the future, we plan to expand our dataset with larger videos and more tasks to increase its impact.


%% file: X_suppl.tex
\clearpage
\setcounter{page}{1}
\setcounter{section}{0}
\renewcommand{\thesection}{\Alph{section}}
\setcounter{figure}{0} 
\setcounter{table}{0} 

\maketitlesupplementary

In this supplementary material, Section A presents SF$^2$T's performance on video caption tasks and additional exemplary visualizations of the attention map, while Section B provides more details about FineVidBench.

\section{More Results and Cases}


In addition to FineVidBench and public video understanding benchmarks, we also evaluated the video caption task (Table~\ref{tab_exp_caption}) using GPT-4o mini, assessing fluency, relevance, informativeness, and correctness, with a maximum score of 40. 
The results show that incorporating SF$^2$T improves performance, highlighting that fine-grained understanding also benefits video captioning. However, after fine-tuning, MiniCPM-V 2.6 produced shorter responses, leading to a decrease in its informativeness score.

\begin{table}[!htbp]
    \renewcommand{\arraystretch}{1}
    \setlength
    \tabcolsep{2.5pt}
    \centering
    \arrayrulecolor{black}
    \footnotesize
    \definecolor{lightgrey}{rgb}{0.95, 0.95, 0.95}
    \begin{tabular}{l|c|c|c|c} 
        \toprule
        \multirow{1}{*}{Methods} & \footnotesize\makecell{LLaVA-NeXT\\-Video~} & \footnotesize\makecell{MiniCPM-V\\2.6~} & \footnotesize\makecell{VideoLLaMA\\2.1~} & \footnotesize\makecell{Qwen2\\-VL~} \\
        \hline
        Base  & 33.20~  & \makebox[2em][r]{32.61~} &  22.53~  & 29.76~ \\
        Base+SF$^2$T & \textbf{33.29}~ & \makebox[3em][r]{29.73~$\downarrow$} & \textbf{30.99}~   & \textbf{30.05}~ \\
        \hline\hline
        Base(SFT)  & 27.62~ & \makebox[2em][r]{29.60~} & 27.19~  & 29.66~ \\
        Base(SFT)+SF$^2$T & \textbf{30.50}~ & \makebox[2em][r]{\textbf{31.31}~} & \textbf{28.94}~   & \textbf{31.04}~ \\
        \bottomrule
    \end{tabular}
    \captionsetup{width=0.48\textwidth}
    \caption{Performance on video caption task. The results show that incorporating SF$^2$T yields higher scores (except MiniCPM-V 2.6), likely due to its enhanced temporal sensitivity and understanding.}
    \label{tab_exp_caption}
\end{table}

\begin{figure*}[!htbp]
    \centering
    \includegraphics[width=0.9\linewidth]{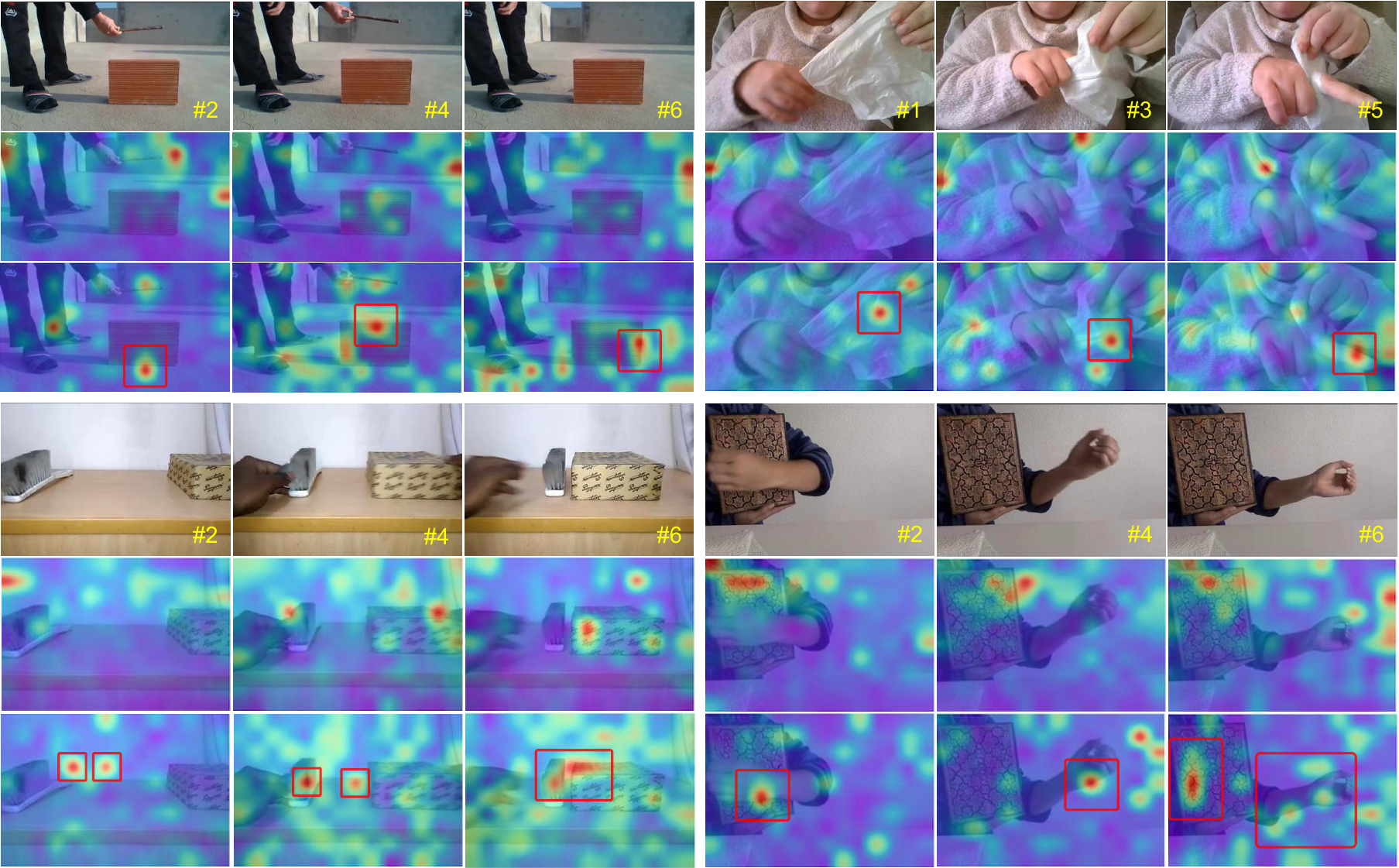}
    \caption{Four exemplary visualizations of the attention map on Qwen2-VL. For each example: top - Original frames; middle - Base (SFT); bottom - SF$^2$T applied. As highlighted by the red boxes, applying SF$^2$T enables the model to better focus on action execution areas and interacting objects, while also predicting the direction of motion.}
    \label{supp_attnmap}
\end{figure*}

As shown in Figure~\ref{supp_attnmap}, we present more attention maps for Qwen2-VL on the Action task, focusing on cases where the model's predictions were corrected after applying SF$^2$T. 

\begin{table*}[!htbp]
    \renewcommand{\arraystretch}{1.5}
    \setlength
    \tabcolsep{3pt}
    \centering
    \arrayrulecolor{black}
    \begin{tabular}{c|c|l} 
        \toprule
        \multicolumn{2}{c|}{Tasks}  & \multicolumn{1}{c}{Question}     \\ 
        \hline
        \multirow{5}{*}{\shortstack{Scene\\Level}}    & Action    & Which activity can be seen in the video?  \\ 
        \arrayrulecolor{black}\cline{2-3}
        & \multirow{3}{*}{Effect}   & After the action takes place, what changes occur to the object?   \\ 
        \cline{3-3}
        &   & During the process of the action, what changes occur to the object?  \\ 
        \cline{3-3}
        &   & After the action takes place, what changes occur in the field of vision?  \\ 
        \cline{2-3}
        & Speed     & What is the rate of movement in the video?   \\ 
        \arrayrulecolor{black}\hline
        \multirow{7}{*}{\shortstack{Fragment\\Level}} & Frame Count  & Could you please tell me how many frames I have inputted?    \\ 
        \arrayrulecolor{black}\cline{2-3}
        & \multirow{3}{*}{Meaning of Order} & \makecell[l]{In the sequence of frames provided, on which frame does the object first appear?}   \\ 
        \cline{3-3}
        &   & \makecell[l]{In the sequence of frames provided, on which frame does the object last appear?}    \\ 
        \cline{3-3}
        &   & In the sequence of frames provided, in which frames does the object exist?   \\ 
        \cline{2-3}
        & Frame Comparison  & Are the two frames I provided exactly the same?   \\ 
        \cline{2-3}
        & Adjust or Not     & \makecell[l]{These frames are all from the same video and capture the dynamic process of an action.\\The order of these frames may have been mixed up. Do we need to rearrange them to \\match the normal execution sequence of the action?}   \\ 
        \cline{2-3}
        & Rearrangement     & \makecell[l]{These frames are all from the same video and depict the dynamic process of an action.\\The order of these frames may have been mixed up. Based on the connections between\\the image frames, which of the following options represents the most appropriate sequence?} \\
        \arrayrulecolor{black}
        \bottomrule
    \end{tabular}
    \caption{Question templates authored by researchers undergo revision by GPT-4o, which rephrases them to maintain the original intent while introducing varied sentence structures and vocabulary.}
    \label{supp_qatemp}
\end{table*}

\section{Details of FinevidBench}
\label{sec:supp_bench}

\subsection{Question-Answer Templates}

Table~\ref{supp_qatemp} delineates the question templates for each task. For the answers, Scene-level tasks include Action task, which are composed of the ``visual synonyms'' and other verbs; Effect task, which are scripted by researchers based on video content; and Speed task, which offer fixed options: fast, slow, normal, and no speed. Fragment-level tasks encompass Frame Count, with answers ranging from 2 to 6; Meaning of Order, using ordinal numbers as responses; Frame Comparison and Adjust or Not, with responses of Yes, No, and Not sure; and Rearrangement, where the answer is a permutation of N numbers, with N representing the number of input frames.
The Question-Answer database is generated through a process of template creation followed by iterative refinement using GPT-4. For Action and Effect tasks, each original video is queried three times using different question formulations. For Speed tasks, one query is conducted for both the original and the speed-altered versions of the video. For Fragment-Level tasks, all five questions are posed for each unique frame count.

\subsection{Detailed Results}

\subsubsection*{$\bullet$ Scene Level}

Table~\ref{supp_effect_class} illustrates the types of action effects and examples in the Effect tasks. For the affected objects, common physical attributes and quantities of objects are considered; notably, the positional relationship, spatial distance, and similarity between two objects are examined. Regarding action attributes, the intensity and completeness of the action are evaluated. Special actions include slight movement, multiple-object movements where several affected objects undergo motion, and compound movements involving two or more atomic actions linked in time. Additionally, camera movements and the inclination of the surface on which objects move are assessed. Table~\ref{supp_effect_class_result} presents the results categorized under the Effect classification. Overall, models performed well in Physical Attributes and Action Intensity, likely due to the ability to infer such information by comparing images before and after the action occurs. However, models exhibited subpar performance in Action Completion and Camera Motion. The former suggests a lack of understanding regarding the distinction between completed and incomplete actions in terms of their effects, while the latter is attributable to the inherent variability and complexity of camera movements. For other tasks, the majority of models exhibited moderate performance. 


\subsubsection*{$\bullet$ Fragment Level}

Table~\ref{supp_varying_frame_count} presents the results for all tasks in the fragment level under varying input frame counts. From the results, we can observe that except for Video-CCAM, the models' ability to count frames significantly declines as the frame count increases. 
Regarding the understanding of order concepts, most models show a clear upward trend, except for ShareGPT4Video.
Models generally perform well on the frame comparison task, likely due to extensive training with image-text pairs. Since the input consistently involves two frames, the results show no significant variation, as expected.
For Rearrangement, all results hover around random values, suggesting that while models recognize incorrect sequence orders, they cannot correct them,  indicating a failure to grasp the dynamic processes of videos truly. 

\begin{table*}[!htbp]
    \renewcommand{\arraystretch}{1.3}
    \setlength\tabcolsep{3pt}
    \centering
    \arrayrulecolor{black}
    \begin{tabular}{c|c|l} 
        \toprule
        \multicolumn{2}{c|}{Effect Type}  & \multicolumn{1}{c}{Examples}   \\ 
        \hline
        \multirow{4}{*}{\shortstack{Object\\Properties}}    &  \begin{tabular}[c]{@{}c@{}}Physical Properties\end{tabular} & \begin{tabular}[c]{@{}l@{}}What~modifications~occur~to~the~\textbf{wafer~stick}~as~a~result~of~the~action?\\A.~Not~sure~~B.~Nothing~happened~~C.~It~broke~~D.~It~deformed\end{tabular}    \\ 
        \arrayrulecolor{black}\cline{2-3}
    & Quantity    & \begin{tabular}[c]{@{}l@{}}Once~the~action~occurs,~what~changes~are~made~to~the~\textbf{mugs}?~\\A.~There~are~about~5~or~6~mugs~here~~B.~There~are~about~1~or~2~mugs~here~~\\C.~There~are~about~3~or~4~mugs~here~~D.~Not~sure\end{tabular}  \\ 
        \arrayrulecolor{black}\hline
        \multirow{9}{*}{\shortstack{Object\\Relationships}} & Position    & \begin{tabular}[c]{@{}l@{}}What~adjustments~take~place~in~the~\textbf{egg}~following~the~action?~\\A.~An~object~appeared~on~top~of~it~~B.~An~object~appeared~in~front~of~it~~\\C.~An~object~appeared~inside~it~~D.~An~object~appeared~behind~it\end{tabular}  \\ 
        \arrayrulecolor{black}\cline{2-3}
    & Distance    & \begin{tabular}[c]{@{}l@{}}What~changes~happen~to~the~\textbf{chili}~and~the~\textbf{cucumber}~after~the~action~is~performed?\\A.~They~grew~more~distant~~B.~It's~unclear~~\\C.~They~came~nearer~D.~Their~separation~remained~consistent\end{tabular}    \\ 
        \cline{2-3}
    & Similarity  & \begin{tabular}[c]{@{}l@{}}What~adjustments~take~place~in~the~\textbf{box}~following~the~action?~\\A.~One~thing~appeared~above~it~~\\B.~Several~things~appeared~above~it,~and~they~looked~different~from~each~other \\C. Not sure~ \\D. Several things appeared above it, and they looked similar to each other\end{tabular}  \\ 
        \arrayrulecolor{black}\hline
        \multirow{6}{*}{\shortstack{Action\\Properties}}    & Intensity   & \begin{tabular}[c]{@{}l@{}}What~alterations~are~observed~in~the\textbf{~paper~cups~}after~the~action~is~taken?\\A.~Not~sure~~B.~It~collapsed~~C.~It~broke~~D.~It~remained~standing\end{tabular}    \\ 
        \arrayrulecolor{black}\cline{2-3}
    & \begin{tabular}[c]{@{}c@{}}Completion\end{tabular}    & \begin{tabular}[c]{@{}l@{}}After~the~action~is~done,~what~modifications~occur~to~the~\textbf{onion}?~\\A.~It~appears~unchanged~from~how~it~was~initially~~\\B.~Something~was~visible~at~the~back~of~it\\C.~An~item~appeared~on~its~surface~~\\D.~Something~was~detected~below~it\end{tabular}    \\ 
        \arrayrulecolor{black}\hline
        \multirow{9}{*}{\shortstack{Special\\Actions}}   & Slight Movement  & \begin{tabular}[c]{@{}l@{}}What~adjustments~take~place~in~the~\textbf{shower~pouf}~during~the~action?~\\A.~I'm~uncertain~~B.~It~dropped~to~the~ground~~C.~It~was~nearly~at~rest~~D.~It~ascended\end{tabular}   \\ 
        \arrayrulecolor{black}\cline{2-3}
    &  \begin{tabular}[c]{@{}c@{}}Mutiple-Object\end{tabular}   & \begin{tabular}[c]{@{}l@{}}What~happens~to~the~\textbf{two~chargers}~while~the~action~is~executed?~\\A.~They~crossed~paths~~B.~They~impacted~each~other~~\\C.~They~proceeded~in~the~same~direction~~D.~It's~unclear\end{tabular}   \\ 
        \cline{2-3}
    & \begin{tabular}[c]{@{}c@{}}Compound\end{tabular}   & \begin{tabular}[c]{@{}l@{}}During~the~process~of~action,~what~modifications~are~observed~in~the~\textbf{plate}?~\\A.~It~fell~after~leaving~the~hand~and~did~not~come~back~~\\B.~It~was~continuously~held~without~any~separation \\C.~It~was~detached~from~the~hand~but~later~reattached~~\\D.~Unclear\end{tabular}   \\ 
        \arrayrulecolor{black}\hline
        \multirow{4}{*}{Others}   & \begin{tabular}[c]{@{}c@{}}Camera movement\end{tabular}   & \begin{tabular}[c]{@{}l@{}}What~alterations~are~evident~in~the~\textbf{flower}~while~the~action~is~carried~out?\\A.~It~appeared~to~move~to~the~right~in~view~~B.~It~appeared~to~ascend~in~view~~\\C.~It~appeared~to~move~to~the~left~in~view~~D.~I~can't~determine\end{tabular}  \\ 
        \arrayrulecolor{black}\cline{2-3}
    & \begin{tabular}[c]{@{}c@{}}Surface Inclination\end{tabular} & \begin{tabular}[c]{@{}l@{}}After~the~action~is~taken,~what~changes~are~noticed~in~the~\textbf{cup}?~\\A.~It~was~stationary~on~a~tilted~surface~~B.~It~was~stationary~on~a~horizontal~surface\\C.~Not~sure~~D.~It~rolled~down~a~sloped~surface\end{tabular}  \\
        \arrayrulecolor{black}
        \bottomrule
    \end{tabular}
    \caption{Types of Effect Task}
    \label{supp_effect_class}
\end{table*}

\begin{table*}[hp]
    \renewcommand{\arraystretch}{1.4}
    \setlength\tabcolsep{3pt}
    \centering
    \arrayrulecolor{black}
    \begin{tabular}{c|c|c|c|c|c|c|c|c} 
        \toprule
        \multicolumn{2}{c|}{Effect Type (Random: 25.00)}   & \begin{tabular}[c]{@{}c@{}}LLaVA-\\NeXT-Video\end{tabular} & \begin{tabular}[c]{@{}c@{}}MiniCPM\\-V 2.6\end{tabular} & \begin{tabular}[c]{@{}c@{}}Video\\LLaMA 2.1\end{tabular} & Qwen2-VL  & \begin{tabular}[c]{@{}c@{}}ShareGPT4-\\Video\end{tabular} & \begin{tabular}[c]{@{}c@{}}Video-\\CCAM\end{tabular}   & \begin{tabular}[c]{@{}c@{}}Avg.\end{tabular}  \\ 
        \hline
        \multirow{2}{*}{\begin{tabular}[c]{@{}c@{}}Object\\Properties\end{tabular}}    & Physical~Properties & 44.20~  & 49.28~   & 52.17~   & \uline{60.87}~    & 47.54~    & \textcolor{red}{63.48}~    & 52.92~ \\ 
        \arrayrulecolor{black}\cline{2-9}
    & Quantity   & 33.33~  & 47.62~   & 56.19~   & \uline{58.10}~   & 41.90~    & \textcolor{red}{60.95}~   & 49.68~ \\ 
        \arrayrulecolor{black}\hline
        \multirow{3}{*}{\begin{tabular}[c]{@{}c@{}}Object\\Relationships\end{tabular}} & Position   & 41.03~  & \uline{51.28}~  & 49.23~   & \textcolor{red}{54.36}~ & 40.31~    & 50.36~    & 47.76~  \\ 
        \arrayrulecolor{black}\cline{2-9}
    & Distance   & 39.56~  & \uline{46.67}~   & 40.89~   & 40.44~ & 40.44~    & \textcolor{red}{48.44}~   & 42.74~  \\ 
        \cline{2-9}
    & Similarity    & 42.86~  & 49.52~   & 47.62~   & \uline{52.38}~   & 38.10~    & \textcolor{red}{59.05~}    & 48.25~  \\ 
        \arrayrulecolor{black}\hline
        \multirow{2}{*}{\begin{tabular}[c]{@{}c@{}}Action\\Properties\end{tabular}}    & Intensity  & 40.27~  & 50.67~   & 53.33~   & \uline{61.33}~ & 52.53~    & \textcolor{red}{62.13}~   & 53.38~  \\ 
        \arrayrulecolor{black}\cline{2-9}
    & Completion    & 39.31~  & \uline{43.68}~  & 38.85~   & 35.63~    & \textcolor{red}{48.05}~  & 34.02~   & 39.92~  \\ 
        \arrayrulecolor{black}\hline
        \multirow{3}{*}{\begin{tabular}[c]{@{}c@{}}Special\\Actions\end{tabular}}   & Slight Movement  & 47.92~  & 43.75~ & 41.67~   & \textcolor{red}{72.92}~    & 35.42~    & \uline{54.58}~   & 49.38~  \\ 
        \arrayrulecolor{black}\cline{2-9}
    & Multiple-Object  & 50.00~  & 60.67~  & \textcolor{red}{76.67}~   & \uline{66.67}~ & 40.67~    & 58.67~   & 58.89~   \\ 
        \cline{2-9}
    & Compound   & 48.15~  & 44.44~   & 51.11~   & \uline{52.59}~    & 35.56~    & \textcolor{red}{53.33}~    & 47.53~  \\ 
        \arrayrulecolor{black}\hline
        \multirow{2}{*}{Others}  & Camera~Movement  & \textcolor{red}{33.33}~   & 22.22~  & 28.89~    & 26.67~   & \uline{32.22}~    & 28.89~   & 28.70~  \\ 
        \arrayrulecolor{black}\cline{2-9}
    & Surface Inclination & 28.57~  & 49.52~   & \uline{58.57}~   & \textcolor{red}{60.48}~    & 41.43~    & 51.43~   & 48.33~  \\
        \arrayrulecolor{black}
        \bottomrule
    \end{tabular}
    \caption{The results of the Effect task, dissected into more granular categories. Overall, Qwen2-VL achieved the best results, with Video-CCAM closely following. Notably, models exhibit suboptimal performance in distinguishing completed from incomplete actions, indicating a lack of ability to associate actions with the resulting state changes of objects.}
    \label{supp_effect_class_result}
\end{table*}

\begin{table*}[hp]
    \renewcommand{\arraystretch}{1.2}
    \setlength\tabcolsep{2pt}
    \centering
    \begin{tabular}{c|c|c|c|c|c|c|c|c} 
        \toprule
        \multicolumn{2}{c|}{Input}    & {\cellcolor[rgb]{0.851,0.851,0.851}}(Random) & \multicolumn{1}{r|}{LLaVA-NeXT-Video}   & \multicolumn{1}{r|}{MiniCPM-V 2.6}   & \multicolumn{1}{r|}{VideoLLaMA 2.1}  & \multicolumn{1}{r|}{Qwen2-VL}  & \multicolumn{1}{r|}{ShareGPT4Video}  & \multicolumn{1}{r}{Video-CCAM}  \\ 
        \hline
        \multirow{5}{*}{3} & {\cellcolor[rgb]{0.949,0.949,0.949}}q1 & {\cellcolor[rgb]{0.851,0.851,0.851}}25.00~   & {\cellcolor[rgb]{0.949,0.949,0.949}}20.33~ & {\cellcolor[rgb]{0.949,0.949,0.949}}93.82~ & {\cellcolor[rgb]{0.949,0.949,0.949}}42.86~ & {\cellcolor[rgb]{0.949,0.949,0.949}}97.25~ & {\cellcolor[rgb]{0.949,0.949,0.949}}60.99~ & {\cellcolor[rgb]{0.949,0.949,0.949}}14.18~  \\
    & {\cellcolor[rgb]{1,0.949,0.8}}q2    & {\cellcolor[rgb]{0.851,0.851,0.851}}25.00~   & {\cellcolor[rgb]{1,0.949,0.8}}19.23~    & {\cellcolor[rgb]{1,0.949,0.8}}48.90~    & {\cellcolor[rgb]{1,0.949,0.8}}35.71~    & {\cellcolor[rgb]{1,0.949,0.8}}29.12~    & {\cellcolor[rgb]{1,0.949,0.8}}76.15~    & {\cellcolor[rgb]{1,0.949,0.8}}38.35~  \\
    & {\cellcolor[rgb]{0.867,0.922,0.969}}q3 & {\cellcolor[rgb]{0.851,0.851,0.851}}33.33~   & {\cellcolor[rgb]{0.867,0.922,0.969}}46.96~ & {\cellcolor[rgb]{0.867,0.922,0.969}}80.66~ & {\cellcolor[rgb]{0.867,0.922,0.969}}71.27~ & {\cellcolor[rgb]{0.867,0.922,0.969}}71.82~ & {\cellcolor[rgb]{0.867,0.922,0.969}}88.41~ & {\cellcolor[rgb]{0.867,0.922,0.969}}66.34~  \\
    & {\cellcolor[rgb]{0.886,0.937,0.855}}q4 & {\cellcolor[rgb]{0.851,0.851,0.851}}33.33~   & {\cellcolor[rgb]{0.886,0.937,0.855}}69.23~ & {\cellcolor[rgb]{0.886,0.937,0.855}}65.38~ & {\cellcolor[rgb]{0.886,0.937,0.855}}81.54~ & {\cellcolor[rgb]{0.886,0.937,0.855}}80.00~ & {\cellcolor[rgb]{0.886,0.937,0.855}}75.55~ & {\cellcolor[rgb]{0.886,0.937,0.855}}80.06~  \\
    & {\cellcolor[rgb]{0.988,0.894,0.839}}q5 & {\cellcolor[rgb]{0.851,0.851,0.851}}25.00~   & {\cellcolor[rgb]{0.988,0.894,0.839}}23.85~ & {\cellcolor[rgb]{0.988,0.894,0.839}}23.08~ & {\cellcolor[rgb]{0.988,0.894,0.839}}33.08~ & {\cellcolor[rgb]{0.988,0.894,0.839}}27.69~ & {\cellcolor[rgb]{0.988,0.894,0.839}}23.68~ & {\cellcolor[rgb]{0.988,0.894,0.839}}23.36~  \\ 
        \hline
        \multirow{5}{*}{4} & {\cellcolor[rgb]{0.949,0.949,0.949}}q1 & {\cellcolor[rgb]{0.851,0.851,0.851}}25.00~   & {\cellcolor[rgb]{0.949,0.949,0.949}}19.77~ & {\cellcolor[rgb]{0.949,0.949,0.949}}90.66~ & {\cellcolor[rgb]{0.949,0.949,0.949}}39.89~ & {\cellcolor[rgb]{0.949,0.949,0.949}}96.63~ & {\cellcolor[rgb]{0.949,0.949,0.949}}16.78~ & {\cellcolor[rgb]{0.949,0.949,0.949}}8.96~   \\
    & {\cellcolor[rgb]{1,0.949,0.8}}q2    & {\cellcolor[rgb]{0.851,0.851,0.851}}25.00~   & {\cellcolor[rgb]{1,0.949,0.8}}24.16~    & {\cellcolor[rgb]{1,0.949,0.8}}60.67~    & {\cellcolor[rgb]{1,0.949,0.8}}41.01~    & {\cellcolor[rgb]{1,0.949,0.8}}33.15~    & {\cellcolor[rgb]{1,0.949,0.8}}65.42~    & {\cellcolor[rgb]{1,0.949,0.8}}43.65~  \\
    & {\cellcolor[rgb]{0.867,0.922,0.969}}q3 & {\cellcolor[rgb]{0.851,0.851,0.851}}33.33~   & {\cellcolor[rgb]{0.867,0.922,0.969}}58.76~ & {\cellcolor[rgb]{0.867,0.922,0.969}}78.53~ & {\cellcolor[rgb]{0.867,0.922,0.969}}76.84~ & {\cellcolor[rgb]{0.867,0.922,0.969}}77.40~ & {\cellcolor[rgb]{0.867,0.922,0.969}}87.23~ & {\cellcolor[rgb]{0.867,0.922,0.969}}63.63~  \\
    & {\cellcolor[rgb]{0.886,0.937,0.855}}q4 & {\cellcolor[rgb]{0.851,0.851,0.851}}33.33~   & {\cellcolor[rgb]{0.886,0.937,0.855}}74.42~ & {\cellcolor[rgb]{0.886,0.937,0.855}}79.85~ & {\cellcolor[rgb]{0.886,0.937,0.855}}93.80~ & {\cellcolor[rgb]{0.886,0.937,0.855}}95.35~ & {\cellcolor[rgb]{0.886,0.937,0.855}}87.50~ & {\cellcolor[rgb]{0.886,0.937,0.855}}94.46~  \\
    & {\cellcolor[rgb]{0.988,0.894,0.839}}q5 & {\cellcolor[rgb]{0.851,0.851,0.851}}25.00~   & {\cellcolor[rgb]{0.988,0.894,0.839}}19.38~ & {\cellcolor[rgb]{0.988,0.894,0.839}}14.73~ & {\cellcolor[rgb]{0.988,0.894,0.839}}24.81~ & {\cellcolor[rgb]{0.988,0.894,0.839}}20.93~ & {\cellcolor[rgb]{0.988,0.894,0.839}}23.10~ & {\cellcolor[rgb]{0.988,0.894,0.839}}22.94~  \\ 
        \hline
        \multirow{5}{*}{5} & {\cellcolor[rgb]{0.949,0.949,0.949}}q1 & {\cellcolor[rgb]{0.851,0.851,0.851}}25.00~   & {\cellcolor[rgb]{0.949,0.949,0.949}}17.98~  & {\cellcolor[rgb]{0.949,0.949,0.949}}86.44~  & {\cellcolor[rgb]{0.949,0.949,0.949}}7.45~ & {\cellcolor[rgb]{0.949,0.949,0.949}}96.05~  & {\cellcolor[rgb]{0.949,0.949,0.949}}0.00~  & {\cellcolor[rgb]{0.949,0.949,0.949}}47.61~  \\
    & {\cellcolor[rgb]{1,0.949,0.8}}q2    & {\cellcolor[rgb]{0.851,0.851,0.851}}25.00~   & {\cellcolor[rgb]{1,0.949,0.8}}28.81~    & {\cellcolor[rgb]{1,0.949,0.8}}59.89~    & {\cellcolor[rgb]{1,0.949,0.8}}50.28~    & {\cellcolor[rgb]{1,0.949,0.8}}37.85~    & {\cellcolor[rgb]{1,0.949,0.8}}41.00~    & {\cellcolor[rgb]{1,0.949,0.8}}55.24~  \\
    & {\cellcolor[rgb]{0.867,0.922,0.969}}q3 & {\cellcolor[rgb]{0.851,0.851,0.851}}33.33~   & {\cellcolor[rgb]{0.867,0.922,0.969}}55.68~ & {\cellcolor[rgb]{0.867,0.922,0.969}}67.61~ & {\cellcolor[rgb]{0.867,0.922,0.969}}80.11~ & {\cellcolor[rgb]{0.867,0.922,0.969}}74.43~ & {\cellcolor[rgb]{0.867,0.922,0.969}}89.69~ & {\cellcolor[rgb]{0.867,0.922,0.969}}64.83~  \\
    & {\cellcolor[rgb]{0.886,0.937,0.855}}q4 & {\cellcolor[rgb]{0.851,0.851,0.851}}33.33~   & {\cellcolor[rgb]{0.886,0.937,0.855}}82.81~ & {\cellcolor[rgb]{0.886,0.937,0.855}}84.38~ & {\cellcolor[rgb]{0.886,0.937,0.855}}94.53~ & {\cellcolor[rgb]{0.886,0.937,0.855}}96.88~ & {\cellcolor[rgb]{0.886,0.937,0.855}}91.55~ & {\cellcolor[rgb]{0.886,0.937,0.855}}96.49~  \\
    & {\cellcolor[rgb]{0.988,0.894,0.839}}q5 & {\cellcolor[rgb]{0.851,0.851,0.851}}25.00~   & {\cellcolor[rgb]{0.988,0.894,0.839}}18.75~ & {\cellcolor[rgb]{0.988,0.894,0.839}}16.41~ & {\cellcolor[rgb]{0.988,0.894,0.839}}22.66~ & {\cellcolor[rgb]{0.988,0.894,0.839}}18.75~ & {\cellcolor[rgb]{0.988,0.894,0.839}}23.29~ & {\cellcolor[rgb]{0.988,0.894,0.839}}23.92~  \\
        \bottomrule
    \end{tabular}
    \caption{The results of all tasks in Fragment-Level under varying input frame counts. Questions q1 through q5 correspond to Frame Count, Meaning of Order, Frame Comparison, Adjust or Not, and Rearrangement, respectively.}
    \label{supp_varying_frame_count}
\end{table*}
